\documentclass[letterpaper, 10 pt, conference]{ieeeconf}  

\IEEEoverridecommandlockouts                              

\title{\LARGE \bf
Learning to Walk by Steering:\\Perceptive Quadrupedal Locomotion in Dynamic Environments
}

\author{Mingyo Seo$^{1}$, Ryan Gupta$^{1}$, Yifeng Zhu$^{1}$, Alexy Skoutnev$^{2}$, Luis Sentis$^{1}$, and Yuke Zhu$^{1}$
\thanks{$^{1}$The University of Texas at Austin, $^{2}$Vanderbilt University.}
\thanks{Correspondance: {\tt\small mingyo@utexas.edu}}%
}

\let\labelindent\relax
\DeclareUnicodeCharacter{0301}{\'{e}}
\usepackage{times}
\usepackage{xcolor}

\usepackage{amsfonts}
\usepackage{amsmath}
\usepackage{amssymb}

\usepackage{graphicx}

\usepackage{booktabs}
\usepackage{hyperref}
\usepackage{dsfont}

\usepackage[font=small,labelfont=bf]{caption}
\usepackage{enumitem}
\usepackage[backend=biber,
            hyperref=true,
            url=false,
            isbn=false,
            doi=false,
            backref=false,
            style=ieee,
            citestyle=numeric-comp,
            sorting=nyt,
            block=none]{biblatex}
\usepackage[font=footnotesize]{caption}

\usepackage{algpseudocode}
\usepackage{algorithm}

\setlist[itemize]{leftmargin=*}
\newcommand{\loosepar}{\looseness=-1}

\renewcommand{\bibfont}{\small}
\newcommand{\ourmethod}{{\textsc{PRELUDE}}}
\newcommand{\heuristic}{{DWA}}
\newcommand{\rlnav}{{Hierarchical RL}}

\newcommand{\bcrnnmpc}{{PRELUDE (MPC)}}
\newcommand{\easy}{Easy}
\newcommand{\medium}{Medium}
\newcommand{\difficult}{Hard}

\begin{document}

\maketitle
\thispagestyle{empty}
\pagestyle{empty}

\begin{abstract}
We tackle the problem of perceptive locomotion in dynamic environments. In this problem, a quadrupedal robot must exhibit robust and agile walking behaviors in response to environmental clutter and moving obstacles. We present a hierarchical learning framework, named \ourmethod{}, which decomposes the problem of perceptive locomotion into high-level decision-making to predict navigation commands and low-level gait generation to realize the target commands. In this framework, we train the high-level navigation controller with imitation learning on human demonstrations collected on a steerable cart and the low-level gait controller with reinforcement learning (RL). Therefore, our method can acquire complex navigation behaviors from human supervision and discover versatile gaits from trial and error. We demonstrate the effectiveness of our approach in simulation and with hardware experiments. Videos and code can be found at the project page: \url{https://ut-austin-rpl.github.io/PRELUDE}.
\end{abstract}


\section{Introduction}




Legged locomotion in natural environments is an essential step toward unlocking real-world robotic applications, such as autonomous inspection, disaster response, and last-mile delivery. Although recent years have witnessed great strides in low-cost and reliable quadruped hardware, overcoming the large state-action space associated with legged robots in complex environments remains an open challenge.
Robots must simultaneously perceive surroundings from onboard sensing and perform agile locomotion to facilitate this advancement. 
%
%
In recent years, a burgeoning body of literature has sought to employ data-driven approaches to improve the robustness and versatility of quadrupedal locomotion.
Reinforcement learning, in particular, has led to promising results, including agile behaviors such as balancing, running, jumping, and robust walking, even in the presence of environment uncertainty~\cite{bledt2017policy, lee2020learning, xie2021glide}. 
However, these methods primarily focus on proprioceptive information while ignoring environment sensing, which hinders their applicability in unstructured human environments.

Pioneering work on perceptive locomotion attempted to incorporate onboard sensory data for motion generation~\cite{bellicoso2018advances, buchanan2020perceptive, fankhauser2014robot, focchi2017high, kim2020vision, pongas2007robust, zico2011stanford}. These methods estimate 3D environment representations (e.g., height maps or voxel grids) for downstream controllers. These representations fall short in capturing rich visual information of semantics and dynamics and instead focus on simple, static environments. Deep reinforcement learning (DRL) approaches~\cite {yang2021learning, imai2021vision} aim at training end-to-end policies that directly map sensory data to motor commands in simulation.
In practice, these approaches are limited by the sample complexity of DRL, which is further exacerbated in cluttered and dynamic human environments.
Furthermore, due to a wide reality gap in rendering RGB images combined with the difficulty of simulating realistic human walking behaviors, these approaches are impractical for real-world deployment.


The goal of this work is to develop a perceptive legged locomotion framework that operates in real-world environments among humans and clutter. 
Our key idea is to decompose the locomotion problem into a high-level navigation and a low-level gait generation policy to realize target commands. This decomposition allows us to harness the complementary strengths of two different learning mechanisms: 1) we train the high-level navigation controller with \emph{imitation learning}. By learning from human demonstration, the model acquires complex navigation behaviors through moving crowds and dense clutter. Furthermore, imitation learning on real-world demonstration data eliminates the aforementioned reality gap; and 2) we train the low-level gait controller with goal-conditioned \emph{reinforcement learning}, giving flexibility in discovering versatile gait styles which may be difficult for conventional controllers.


\begin{figure}[!t]
	\centering
	\includegraphics[width=\linewidth]{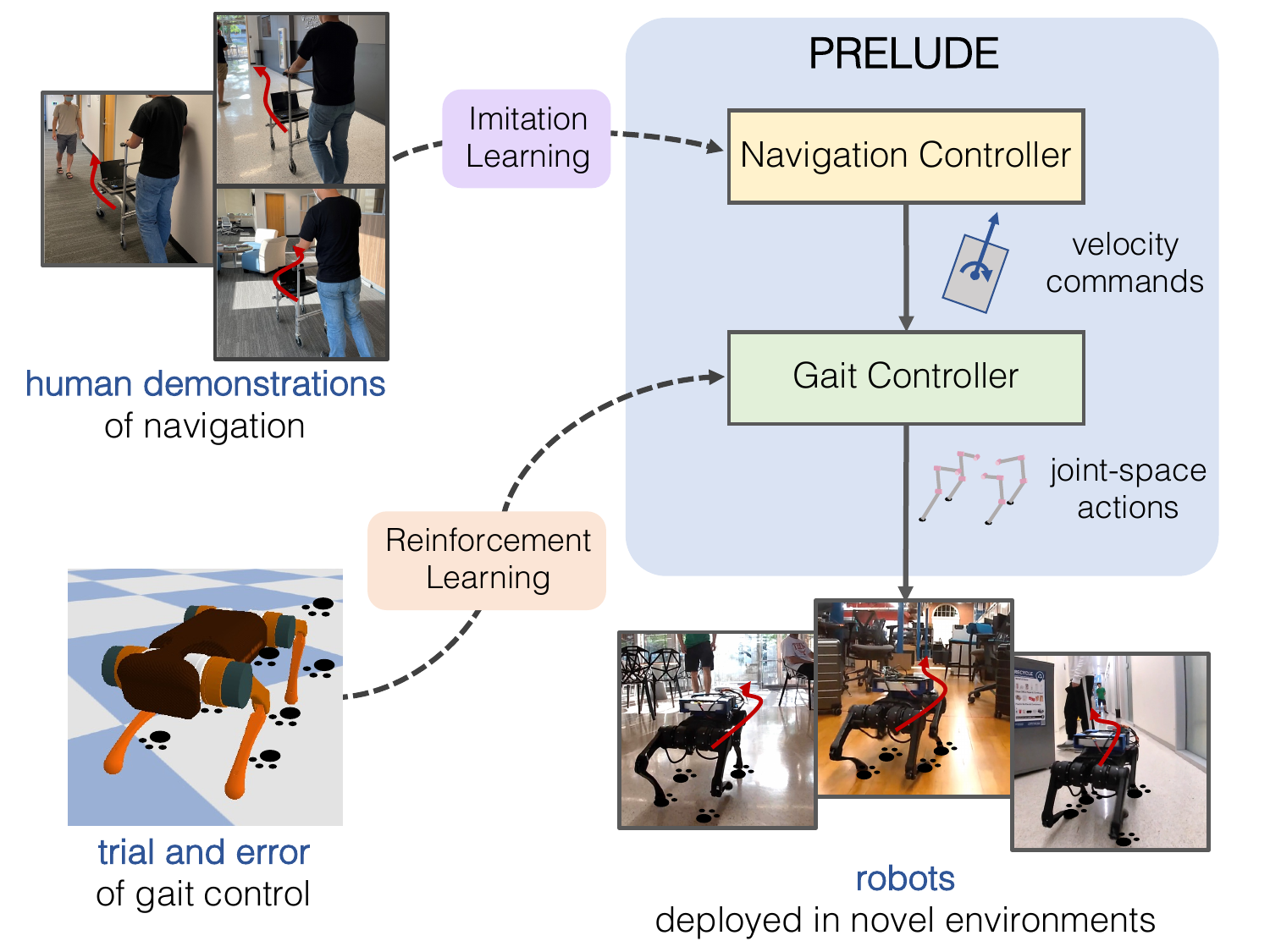}
	\caption
	{\textbf{Method overview.} \ourmethod{} tackles the problem of perceptive locomotion in dynamic environments. We introduce a control hierarchy where the high-level controller, trained with imitation learning, sets navigation commands and the low-level gait controller, trained with reinforcement learning, realizes the target commands through joint-space actuation. This combination enables us to effectively deploy the entire hierarchy on quadrupedal robots in real-world environments.
	}
	\label{fig:pull-figure}
\end{figure}


We name our method \ourmethod{} (\underline{P}e\underline{r}ceptiv\underline{e} \underline{L}ocomotion \underline{U}nder \underline{D}ynamic \underline{E}nvironments). It consists of a high-level navigation controller and a low-level gait controller (see Figure~\ref{fig:pull-figure}). 
Both controllers are implemented as deep neural networks. 
The navigation controller, trained with imitation learning, takes in egocentric RGB-D data 
and generates a desired motion command. 
To collect data for imitation learning, we design an intuitive steerable cart that can be driven around to collect steering actions in natural environments. A key advantage of this design is that a non-expert human can easily supply demonstrations at scale by pushing the cart around. 
The navigation controller, trained on data collected on the cart, is able to leverage rich human supervision to perform complex spatiotemporal reasoning and generate dynamic robot motions. 
Meanwhile, the gait controller, trained with reinforcement learning, learns to achieve variable velocity commands as input. During deployment, it takes as input the target motion command given by the high-level navigation controller and realizes the desired motion. 

We evaluate our approach in simulation and with real hardware. In simulation, we demonstrate that our method outperforms state-of-the-art reinforcement learning (RL) methods by 35.4\%. 
Further, we illustrate that our RL locomotion controller exhibits superior 
performance over conventional model-based methods. 
The hierarchical design of our method allows us to train the locomotion controller in large-scale simulation and readily transfer it to the real world. 
Meanwhile, the navigation controller can be trained directly from real-world human demonstration data. By combining these two facets, we can effortlessly deploy the real hardware in a realistic environment.



Our work has three main contributions: 1) We introduce \ourmethod{, }a hierarchical model for perceptive locomotion in dynamic environments and develop effective strategies to combine human supervision and reinforcement learning;
2) We develop a practical cart design to collect human demonstration data at scale for training our model;
3) We show the effectiveness of our method in simulation and on physical hardware and perform ablative studies to analyze individual model components. We show \ourmethod{} generalizes to unseen real-world environments with diverse objects and human behaviors.
\section{Related Work}

\paragraph{Quadrupedal Locomotion} 
The control of legged robots has been studied for decades. 
One of the main challenges is high computational requirements for optimizing whole-body dynamics.
In the last decade, the centrifugal model, which considers the robot as a lumped mass to reduce the computational burden, has been widely used by optimization-based methods~\cite{winkler2018gait, bledt2019implementing, kim2019highly, carius2020mpc}.
However, such model-based approaches are constrained by computational limits and model mismatch.
In response to these limits, the integration of model-based methods and data-driven methods has been actively studied \cite{da2021learning, yang2022fast, xie2021glide, rudin2022learning}.
These methods leverage knowledge of robot dynamics encapsulated in model-based methods to improve training efficiency.
However, the boost in training efficiency is hindered by the computational costs of the underlying controller. 
%
Alternatively, there has been some recent success when training end-to-end policies for locomotion \cite{kumar2021rma, fu2021minimizing}.
These policies boast fast computation speeds when inferring actions and have been shown to outperform model-based methods. 
However, these approaches focus on improving legged robots' 
agility rather than how legged locomotion is applied to real tasks.


\paragraph{Perceptive Locomotion of Quadrupedal Robots}
Legged robots with exteroceptive sensors have been studied for decades. 
Several model-based approaches exploit exteroceptive sensory modules to reconstruct the shape of surrounding terrains~\cite{zico2011stanford, belter2016rgb, buchanan2020perceptive, magana2019fast, jenelten2020perceptive, kim2020vision, gangapurwala2022rloc}.
However, they rely on accurate terrain reconstruction, and the locomotion performance is limited by the precision and speed of reconstruction.
Another family of work has employed neural network models to transform raw visual observations into navigation commands~\cite{sorokin2022learning, pan2020zero, hoeller2021learning}. However, they do not account for robot dynamics and instead use a pre-defined gait controller, which is not optimized for the target tasks.
There is a growing interest in end-to-end learning of perceptive locomotion~\cite {yang2021learning, imai2021vision}. These methods require meticulous simulation training and sim-to-real transfer, which largely limits their applicability to (quasi-)static environments.
Inspired by recent advances, our approach uses deep learning models to tackle this problem. Rather than seeking an end-to-end algorithm, we design our method to harness the complementary strengths of imitation and reinforcement learning, allowing the learned model to be deployed to real hardware with minimal domain gaps.

\paragraph{Hierarchical Frameworks for Locomotion}
Hierarchical frameworks have been used in both model-based and data-driven locomotion.
The main advantage of hierarchical frameworks is their ability to break down complicated systems into multiple modules, such that each module can be designed with domain-specific knowledge~\cite{kim2019highly, bledt2019implementing, carius2020mpc}.
In particular, hierarchical approaches to motion generation offers several advantages when using data-driven methods. 
First, low-level control can be reused with various types of high-level gait planners and vice versa. 
Second, the overall system can be separated into modules, and individual modules are easier to train and analyze~\cite{peng2017deeploco, jain2019hierarchical, tsounis2020deepgait}.
Our approach also adopts a hierarchical structure. We demonstrate how it gives rise to a practical learning algorithm for perceptive locomotion in dynamic environments.

\section{Method}

\begin{figure*}[!t]
	\centering
\vspace{4pt}
\includegraphics[width=\linewidth]{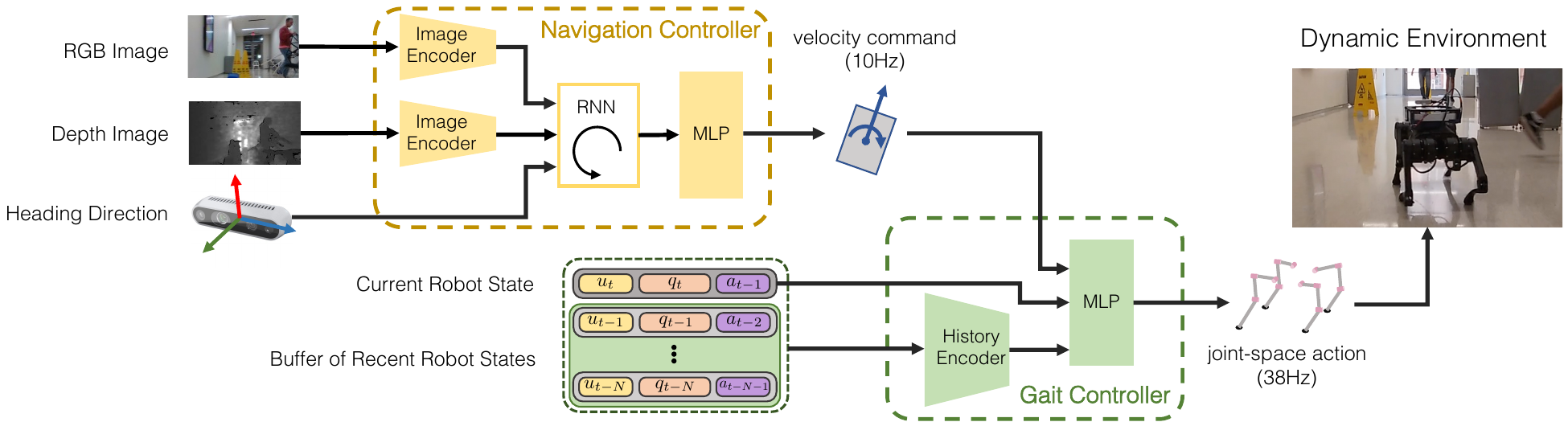}
	\caption
	{
        \textbf{Model architecture of \ourmethod{}.}
        The high-level navigation policy generates the target velocity command $u_t$ at 10Hz from the onboard RGB-D camera observation and robot heading.
        The target velocity command is used as input to the low-level gait controller along with the buffer $\mathcal{B}_t$ of velocity commands ${u}_t$, recent robot states ${q}_t$ and previous joint-space actions ${a}_{t-1}$. 
        The low-level gait policy predicts the joint-space actions as the desired joint positions at 38Hz and sends them to the quadrupedal robot for actuation.
	}
	\label{fig:model}
\end{figure*}

We introduce \ourmethod{}, a hierarchical learning framework for quadrupedal locomotion and navigation in dynamic environments using onboard vision sensing. The key to our approach is to decompose the perceptive locomotion pipeline into a two-level hierarchy consisting of a high-level navigation controller and a low-level gait controller. In this section, we describe our problem formulation and then introduce the hybrid learning scheme used to train the controllers at the two levels of the hierarchy: 1) imitation learning with human supervision for training the navigation controller and 2) reinforcement learning for training the locomotion controller.\loosepar{}

\subsection{Problem Formulation}

We model the problem of perceptive legged locomotion as a discrete-time Markov Decision Process $\mathcal{M}=(\mathcal{S}, \mathcal{A}, \mathcal{P}, R, \gamma, \rho_{0})$ where $\mathcal{S}$ is the state space, $\mathcal{A}$ is the action space, $\mathcal{P}(\cdot|s, a)$ is the stochastic transition probability, $R(s, a, s')$ is the reward function, $\gamma \in [0, 1)$ is the discount factor, and $\rho_{0}(\cdot)$ is the initial state distribution. Our goal is to learn a closed-loop visuomotor policy $\pi(a_t|s_t)$ that maximizes the expected return $\mathbb{E}[\sum^\infty_{t=0}\gamma^t R(s_t, a_t, s_{t+1})]$. In our context, $\mathcal{S}$ is the space of the robot's sensor data captured by its egocentric camera and proprioceptive sensors, $\mathcal{A}$ is the space of the robot's joint-space commands,  $R(s, a, s')$ is the reward function designed for the locomotion task, and $\pi$ is a closed-loop policy that we deploy on the robot. 

To handle the complexity of perceptive locomotion and train the policy $\pi$ effectively, we decompose the policy $\pi$ into a two-level hierarchy. At the high level is a navigation policy $\pi_{H}$ that computes the target motion command $u \in \mathbb{R}^2$ as the target linear and angular velocities. At the low level is a gait policy $\pi_L$ that computes the robot's joint-space actions to realize target command $u$ from $\pi_H$. With this hierarchical abstraction, we can rewrite the whole policy as, $\pi(a_t|s_t) = \pi_{L}(a_t|s_t, u_t)\pi_{H}(u_t|s_t) $.
\loosepar{}

\subsection{Hierarchical Perceptive Locomotion Policy}

\ourmethod{} consists of a high-level policy $\pi_H$ for predicting navigation commands and a low-level $\pi_L$ for generating locomotion gaits (see Figure~\ref{fig:model}).
Here we describe the scheme we used to train both policies and their seamless deployment both in simulation and on the real robotic system.
\loosepar{}

\paragraph{Imitating Navigation Demonstrations in Dynamic Environments} 
When surrounded by static obstacles and people moving in and out of the scene, the robot must anticipate future motion and subsequently generate high-quality motion plans.
By leveraging imitation learning, \ourmethod{} avoids the difficulty of training an RL policy that must simulate full human walking dynamics and also attempt effective navigation.
Instead, we propose to learn the policy $\pi_H$ for navigation using imitation learning. Through imitating human navigation behaviors, the robot is able to acquire the ability of complex spatial reasoning and motion prediction in dynamic environments. Meanwhile, this form of high-level demonstration is easy to collect on a hardware platform of a much simpler morphology than the quadruped. We describe our hardware platform of the steerable cart in Sec.~\ref{sec:exp-setup}.

The collected demonstration dataset $\mathcal{D}$ consists of state-action pairs $\mathcal{D}=\{(s_i, u_i)\}_{i=1}^N$, where the states are the camera observations in a similar view as of the robot's onboard egocentric camera plus the camera orientation while the actions are the linear and angular velocities estimated by odometry, and $N$ is the total number of data points. With this dataset, we train the navigation policy $\pi_H$ using a supervised imitation learning algorithm. Specifically, we design a Behavioral Cloning model with recurrent neural networks (RNNs)~\cite{mandlekar2021matters} to capture the temporal information of moving objects in the environments. Furthermore, demonstrations collected from human operators are noisy and multimodal in nature. To capture the multimodality of action distributions while preventing mode collapsing, the policy uses a Gaussian Mixture Model (GMM) to parameterize a distribution of actions~\cite{wang2020critic}.

In practice, $\pi_H$ is implemented as a deep neural network. At each time step, its input consists of the egocentric RGB-D observation and the heading direction of the body. It outputs the target velocity command, i.e., two scalar values for forward-linear velocity and angular velocity. As the RGB images provide rich task-relevant semantic information and the depth images provide geometry information of surrounding environments, we process RGB images and depth images separately through individual \textit{image encoders}. The target velocity command serves as a suitable representation for specifying high-level navigation behaviors, as it can define highly reactive motions in dynamic environments compared to other common representations, such as position-based commands.
\loosepar{}

\paragraph{Learning Robust Locomotion Gaits}  
A robust gait controller is key for locomotion in real dynamic environments. To this end, a desirable gait controller should meet the following two requirements: 1) it can track constantly varying velocity commands set by $\pi_H$ to generate agile walking behaviors for dynamic collision avoidance; 2) it can be effortlessly deployed to real hardware in the presence of dynamics uncertainty. Motivated by recent successes in learning gait controllers in simulation and sim2real transfer~\cite{da2021learning,kumar2021rma,lee2020learning}, we use large-scale reinforcement learning in simulation to train our gait controller. 
Concretely, \ourmethod{} learns the policy using the distributed Proximal Policy Optimization (PPO) algorithm~\cite{schulman2017proximal} in parallel copies of simulated environments. We randomize the target velocity commands (as input to the gait policy $\pi_L$) to learn a gait policy to track rapidly changing commands. The policy is optimized over the reward function designed for 1) minimizing the tracking command error and energy, 2) stabilizing the robot body, and 3) encouraging concrete foot contacts. In addition, we introduce two key designs to ensure the learned policy to transfer directly to physical hardware: First, we apply aggressive \textit{domain randomization}~\cite{tan2018sim} to physical parameters in simulation to improve the robustness of our controller while bypassing the complicated system identification process on the real robot. We randomize mass, inertia, friction coefficients of robot joints, and external force perturbation. Second, we use a \textit{buffer of recent robot states} as input to the policy $\pi_L$ rather than raw visual observations. For each time $t$, the input buffer $\mathcal{B}_t$ contains $T+1$ tuples $\mathcal{B}_t=\{(u_i,q_i,a_{i-1})\}_{i=t-T,\ldots,t}$, where $u_i$ is the navigation command at time $i$, $q_i$ is the proprioceptive state at time $i$, and $a_{i-1}$ is the joint-space action output of $\pi_L$ from the previous time step $i-1$. This design resonates with recent work RMA~\cite{kumar2021rma}, which demonstrated that the robot's recent state-action history can serve as a robust proxy for estimating extrinsics. For perceptive locomotion, using the buffer of robot states as input eliminates the needs for us to handle the reality gap of visual observations from camera sensors.
\section{Experiments}
\label{sec:exps}

\begin{figure}[t]
	\centering
\vspace{4pt}
\includegraphics[width=\linewidth]{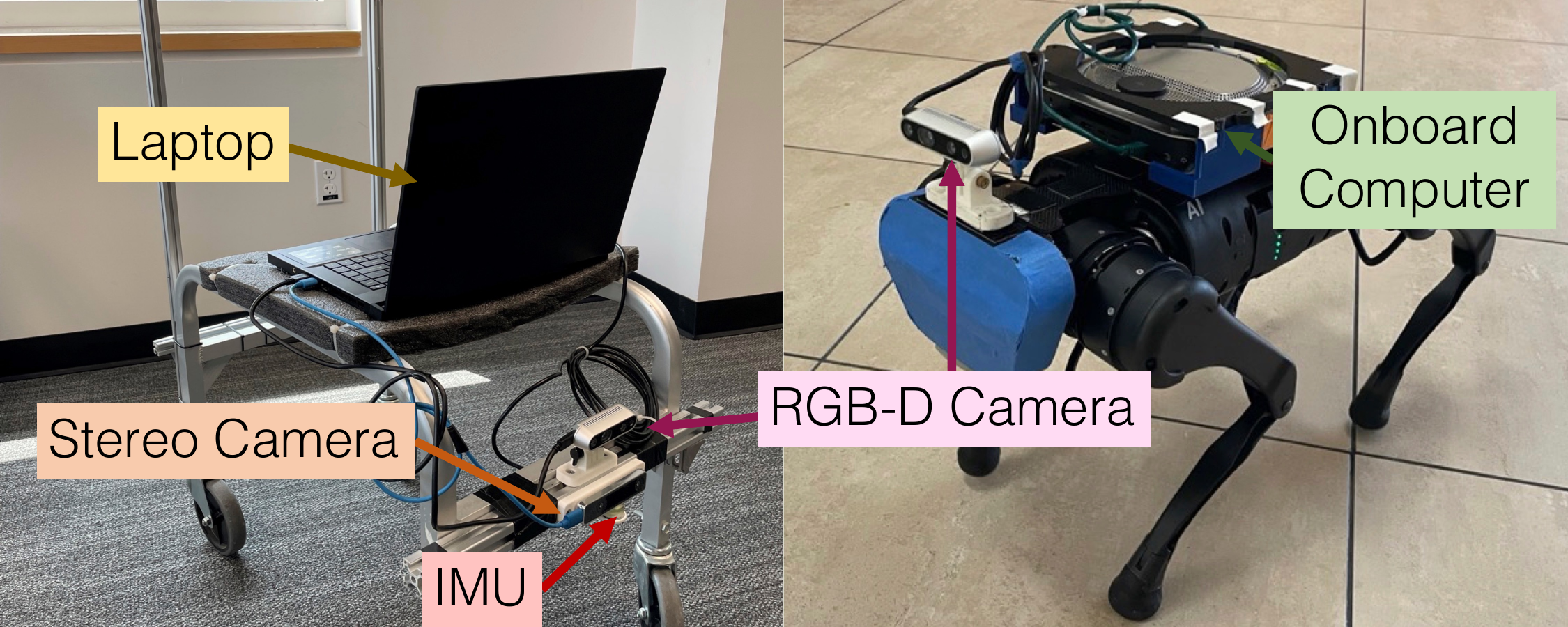}
	\caption
	{
	\textbf{Hardware platforms.} A steerable cart designed for collecting human demonstrations of navigation (left) and the Unitree A1 robot with ego-centric RGB-D camera mounted at cart height (right). It ensures the navigation policy trained on demonstration data can be directly deployed on the robot.
	}
	\label{fig:hardware}
\end{figure}

We design our experiments to answer the following three questions: 1) How does \ourmethod{}'s hierarchical design perform compared to existing methods for perceptive locomotion? 2) Is our learned low-level gait controller more robust than conventional model-based controllers in motion generation? 3) Can \ourmethod{} be deployed on the quadruped robot in real-world environments?
\loosepar{}



\subsection{Experimental Setup}
\label{sec:exp-setup}

We validate our proposed method for perceptive locomotion on a track with static obstacles and people who may be moving around. 
A successful episode is one where the robot walks from one end to the other without falling or colliding with obstacles, walls, or people. 
 We design environments of varying difficulty to analyze our model's performance in simulation and in real hardware with the Unitree A1 quadrupedal robot~\cite{unitree2021}.
\loosepar{}

\paragraph{Simulation Setup}
We develop a simulated corridor environment in Bullet Physics~\cite{coumans2017pybullet} with 3m width and 50m length. We use this environment to systematically compare our approach to baselines under various conditions.
We initialize the scenes with different numbers of static obstacles and walking humans, then categorize the conditions into three difficulty levels:
\begin{itemize}
\item {\bf \easy{} Level:} Corridor environments with sparsely placed obstacles (average 5) and zero or one person moving.
\item {\bf \medium{} Level:} Cluttered environments with on average 15 static obstacles and zero or one person. Due to the increase in objects, the robot is more prone to collision.
\item {\bf \difficult{} Level:} Challenging environments with on average 15 obstacles and 4 humans. The crowd raises the uncertainty, so the robot must anticipate the human motions.
\end{itemize}
All the static obstacles are initialized with 3D assets from the ShapeNet~\cite{shapenet2015} and Google Scanned Objects~\cite{IgnitionFuel-GoogleResearch-Google-Scanned-Objects} that cover everyday objects. 
Human motion is governed by a Gaussian Process-based motion generation method. 
Further, we add random external perturbations to robots to mimic the environmental uncertainties that tend to exist in real-world scenarios.\loosepar{}

\paragraph{Real-World Setup}

We consider our real-world deployment environment as 15m tracks with different numbers of static obstacles and walking humans, then categorize the conditions into four types with two annotations: 
\begin{itemize}
\item {\bf Static (Sparse):} Open space with on average 10 large pieces of furniture with enough space to bypass them. 
\item {\bf Static (Complex):} Narrow office area with more than 10 obstacles where the robot may pass through tight spaces.
\item {\bf Dynamic (Sparse):} Open space with on average 5 humans and 5 static obstacles. The robot has enough space to bypass humans but needs to anticipate humans' movements.
\item {\bf Dynamic (Complex):} Narrow corridor with 1.6m in length with on average 4 humans and less than 5 obstacles, the most challenging environments where the robot has to pass through narrow space and rapidly react to humans.
\end{itemize}
Between each trial, obstacles were shuffled around the area and swapped for others. 
People are instructed to walk in natural ways. 

The deployment environments are different from where human demonstration data are collected, with unseen backgrounds, new objects, unexpected human behaviors, and changing light conditions.
We mount a RealSense D435 \cite{realsense2022} camera on the forehead of the robot for egocentric RGB-D observations. The gait controller predicts desired joint positions at 38Hz, which is transmitted to the internal Unitree SDK for actuation. The controller is deployed on an onboard Apple Mac mini M1 computer for real-time inference.\loosepar{}

\paragraph{Data Collection}

To collect demonstration data using the quadruped would require a human expert to control the robot in complex environments over long periods of time. This approach is hard to scale due to the mechanical limits of the hardware and the difficulty of control. To ease data collection, we design a steerable cart platform (see Figure~\ref{fig:hardware}). 
This cart platform is designed in a non-holonomic morphology, permitting only the forward linear velocity and the angular velocity. Human operators can easily maneuver this cart among static and dynamic obstacles to generate expert data.
We mount an Intel RealSense D435 camera on the cart to record RGB-D observations. 
We further obtain the cart's linear velocity and yaw-direction angular velocity from the front-mounted Intel RealSense T265 stereo camera. The heading direction can be estimated by the VectorNav VN100 IMU sensor installed on the cart. 
This way, we eliminate the need for manual labeling of human actions.
We collect $250$ real-world trajectories within three hours of wall-clock time. To mimic such a steering maneuver in simulation, we use the 3Dconnexion SpaceMouse 
to provide velocity commands. We collect $80$ demonstration trajectories for each difficulty level. For both domains, our datasets consist of the following information: 1) observation data, including RGB-D images and the heading orientation; 2) the forward linear and angular velocity of the cart.
\loosepar{}

\begin{figure*}[t]
	\centering
        \vspace{4pt}
        \includegraphics[width=1.\linewidth]{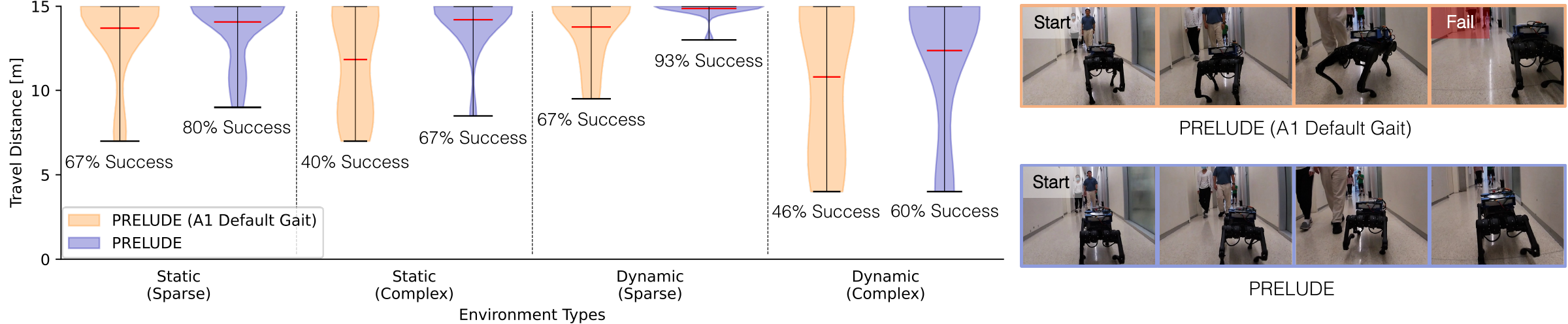}
	\caption
	{
        \textbf{Real robot experiments.} \textbf{(Left)} We perform real-world trials where the robot traverses 15m-length tracks in different configurations. The plot shows distributions of travel distances in meter. The black and red lines indicate the ranges and the means of traverse length, respectively. \textbf{(Right)} We observed that \ourmethod{} (A1 Default Gait) drifts aggressively after a high-speed turning and collides into the wall, while \ourmethod{} turns rapidly to bypass the walking crowd and completes the trial successfully.
	}
	\label{fig:real-robot}
\end{figure*}

\subsection{Quantitative Evaluation}
\label{sec:quant-eval}


\paragraph{Simulation Evaluation} We compare \ourmethod{} against the following baselines:
\label{sec:sim-eval}
\begin{itemize}
\item{\bf Dynamic Window Approach} (\heuristic{}): An online collision avoidance method that computes a cost map from the obstacles' positions and plans navigation commands based on the cost map \cite{fox1997dynamic}. 
It uses the navigation commands with our learned gait controller to actuate the robot. This baseline accesses ground-truth positions of all obstacles in the field of view.
\item{\bf \rlnav{}:} This baseline trains the high-level navigation controller with reinforcement learning (RL) using the same distributed PPO algorithm. It uses the navigation command as the action space and invokes our learned gait controller. The reward is defined as the progress of moving forward at each step. This strategy has been explored in prior work~\cite{peng2017deeploco, xia2020relmogen, nasiriany2022augmenting}, which uses low-level action primitives to level up the action abstraction for RL.
\item{\bf \bcrnnmpc{}:} This baseline uses a conventional Model Predictive Control (MPC) baseline~\cite{di2018dynamic} instead of our learned gait controller. We assume the MPC baseline accesses ground-truth body velocity.
\item{\bf \ourmethod{} (BC):} A variant of our final model, which does not include the recurrent neural networks in the navigation policy architecture. All the other parts remain the same.
\end{itemize}
To quantitatively evaluate each model, we use the success rate of completing the track and the travel distance, averaged across all test trials. The distance metric allows us to measure partial progress made by the baselines when they struggle to complete the entire task. To evaluate all the models consistently and control the environment setup, we generate $50$ randomized scenes for each difficulty level and use the same set of 50 scenes for evaluating each model. We present the average length traversed and success rate in Table~\ref{tab:nav}.


Table~\ref{tab:nav} suggests that \ourmethod{} achieves the overall best performance in all three difficulty levels. Results of the \heuristic{} baseline imply that all three levels of environments require non-trivial collision avoidance behaviors. Meanwhile, the \rlnav{} baseline shows that decoupling gait control and navigation allowed the robot to traverse in simple and static environments. However, RL alone struggles to discover sophisticated navigation strategies, especially for handling dynamic obstacles.
Results of \ourmethod{} (BC) indicate that the lack of temporal modeling leads to inferior performances, especially in the Hard level, where our final model traverses a 44\% longer average distance than the BC baseline. We hypothesize that the temporal information, harnessed by our recurrent policies, plays a critical role in reasoning about moving people and anticipating their future motions.

Results of \bcrnnmpc{} indicated that our learned navigation controller is compatible with different types of gait controllers. Even though \bcrnnmpc{} takes advantage of the robot's dynamic models for MPC, our RL-based gait achieved higher performances overall. These results imply that our learned gait controller exhibits high robustness for agile navigation in dynamic environments. \loosepar{}

\begin{table}[t]
\centering
\vspace{23pt}
\caption{\textbf{Comparison of~\ourmethod{} and baselines in simulation.} We report the average length traversed in meters (total track length: 50m) and success rate in percentage as the evaluation metric.}
\makeatletter\def\@captype{table}
\resizebox{\columnwidth}{!}
{
  \begin{tabular}{lcccc}
    \toprule
    \textbf{Method} & \textbf{\easy} & \textbf{\medium} & \textbf{\difficult}\\
    \midrule 
     DWA~\cite{fox1997dynamic} & {42.2} $\pm$ 12.8 (66\%) & {34.2} $\pm$ 18.2 (52\%) & {30.2} $\pm$ 18.3 (40\%)\\
     \rlnav{}~\cite{peng2017deeploco} & {36.4} $\pm$ 14.8 (44\%) & {24.7} $\pm$ 17.6 (26\%) &{28.4} $\pm$ 18.9 (38\%)\\
     \midrule
     PRELUDE (BC) &  {32.0} $\pm$ 16.0 (32\%) & {28.1}  $\pm$ 15.5 (24\%) &  {24.1} $\pm$ 14.9 (16\%)\\
     PRELUDE (MPC~\cite{di2018dynamic})  &  {42.3} $\pm$ 14.3 (74\%) & {38.9} $\pm$ 15.4 (60\%) & {34.0} $\pm$ 18.1 (50\%)\\
     PRELUDE &  \textbf{44.3 $\pm$ 12.4} (\textbf{80\%}) & \textbf{41.6 $\pm$ 15.2} (\textbf{74\%}) &  \textbf{35.3 $\pm$ 18.2} (\textbf{56\%})\\
    \bottomrule
  \end{tabular}
}
\label{tab:nav}
\end{table}

\begin{table}[t]
\vspace{15pt}
\centering
\caption{{\bf Evaluation of gait controllers.} We report the tracking errors in meters (lower the better) and success rate in percentage (higher the better) on 20 trials of tracking each pre-defined trajectory.
}
\makeatletter\def\@captype{table}
  \resizebox{0.85\linewidth}{!}{  
  \begin{tabular}{lcc}
    \toprule
     &\textbf{MPC~\cite{di2018dynamic}} & \textbf{Gait Controller (Ours)} \\
    \midrule
    $x$-axis Linear & 0.26 $\pm$ 0.03 (50\%)  &  {\bf 0.20} $\pm$ 0.02 ({\bf 100\%}) \\
    $x$-axis Sine & 0.28 $\pm$ 0.05 (60\%) & {\bf 0.24} $\pm$ 0.02 ({\bf 95\%}) \\
    $x$-axis Step & 0.29 $\pm$ 0.04 (45\%)  & {\bf 0.26} $\pm$ 0.04 ({\bf 90\%}) \\
    Sine Trajectory & 0.27 $\pm$  0.04 (55\%) & {\bf 0.23} $\pm$ 0.03 ({\bf 90\%})  \\
    Zig-zag Trajectory  & 0.26 $\pm$ 0.03 (55\%)  & {\bf 0.24} $\pm$ 0.02 ({\bf 85\%}) \\
    \bottomrule
  \end{tabular}
  }
  \label{tb:rl}
\vspace{-15pt}
\end{table}

\paragraph{Analysis of Gait Controllers}
We have shown \ourmethod{} outperforms baseline approaches in the perceptive locomotion task and further analyze the quality of the learned gait controller. 
We send the robot velocity commands computed from the set points of a pre-defined position trajectory.  
At each time step, the velocity command is computed by a simple PD controller based on the next set point and the robot's current position.
We set desired velocity commands with a PD controller because unit testing on the locomotion controllers is performed in a closed-loop manner to mimic the integration with Navigation Controller. Furthermore, realistic navigation tasks require fast convergence speed and stability with respect to velocity error.  

We design trajectories with diverse shapes and curvatures to evaluate the agility of our learned gait controller against the MPC baseline~\cite{di2018dynamic}. 
To evaluate that our gait policy handles aggressive changes in linear velocity, we design the below linear trajectories of different set point distances:
\begin{itemize}
\item {\bf $x$-axis Linear}: 
Linear trajectory of set points with constant distances of 0.7m per second.
\item {\bf $x$-axis Sine}:  
Linear trajectory of set points with distances varying as sine functions in the range of $0.4$ to $1.0$m per second.
\item {\bf $x$-axis Step}: Linear trajectory of set points with distances varying as step functions with values of $0.5$, $0.7$, and $1.0$m per second.
\end{itemize}
To show that our gait policy handles large changes in the yaw rate, we designed the additional trajectories below:
\begin{itemize}
\item {\bf Sine Trajectory}: Sine-like trajectory of set points with constant distances. The yaw direction of the trajectory changes in the range of $0.0$ and $1.4$ rad.
\item {\bf Zig-zag Trajectory}: Zig-zag trajectory of set points with constant distances where the direction of the path is given as step functions, as $-0.4$ to $0.4$ rad.
\end{itemize}
We report two evaluation metrics: 1) the quality of trajectory tracking, computed as the average tracking error between the ground-truth trajectories and the robot's realized trajectories, and 2) the robustness of the controller, computed by evaluating the success rate in tracking the trajectories below the error limit of 1m without failing  within 30 seconds.
Table \ref{tb:rl} presents the results. Our Gait Controller, learned with reinforcement learning, realizes the trajectories more accurately and robustly than the MPC baseline. In particular, we observe that the MPC method regulates body acceleration and changes the body motion very smoothly when the changes to the velocity command are small. However, when commands change more aggressively, simplifications of inertia terms to convexify the MPC cause failure of the robot.
In contrast, we observe that our RL-based controller can safely track aggressively changing commands. 
These attributes are suitable for performing the given task of navigation amongst dynamic obstacles where changes need to be rapid. Furthermore, our controller has a significantly lower lateral drift than MPC when tracking velocity commands. 
\loosepar{}

\paragraph{Real Robot Evaluation} 
Finally, we demonstrate that \ourmethod{} can be deployed on the real robot. We compare it with our self-baseline \ourmethod{} (A1 Default Gait), a variant of our final model, using the robot's default model-based controller instead. The comparison shows the effectiveness of \ourmethod{} in real-world deployment. In particular, 1) our navigation controller can be deployed with other types of gait controllers, and 2) our learned gait controller outperforms model-based controllers for locomotion in cluttered and dynamic environments. 

To quantitatively evaluate each model, we report the success rate of completing the track and the distance that the robot traverses before a collision or reaching the end of the track, averaged across all test trials, as shown in Figure~\ref{fig:real-robot}.
To evaluate the models consistently and control the environment setup, we generate $15$ randomized scenes for each type of scenario.
The result shows that both \ourmethod{} and \ourmethod{} (A1 Default Gait) can navigate robustly in static and dynamic environments. It implies that our navigation controller works well with different gait controllers on real robots. Furthermore, \ourmethod{} achieves overall longer travel distances and higher success rates considering all the four types of environments. 
We note that the default gait controller suffers from drifts at high-speed turning, often resulting in collisions with obstacles on the side. We hypothesize that model errors from the simplified dynamic model and inaccurate state estimation caused these issues. 
Our learned gait controller exhibits less drift when turning at high rates, and the robot shows higher agility in cluttered and dynamic environments.\loosepar{}
\section{Conclusion}
\label{sec:conclusion}
We introduce \ourmethod{}, an effective method for learning perceptive locomotion controllers for quadrupedal robots to traverse real-world dynamic environments. Our method combines the complementary strengths of imitation learning and reinforcement learning through a hierarchical design that decomposes the locomotion problem into high-level navigation and low-level gait generation. We designed a steerable cart platform for collecting human navigation demonstrations in complex scenes.
and used the collected datasets to train the high-level navigation policy. We used large-scale reinforcement learning to train the low-level gait controller in simulation, demonstrating its effectiveness in transferring to the real world and producing robust and versatile motions. 
Our work focuses on flat and indoor environments where human steering actions can be conveniently collected with the wheeled platform.
For future work, we would like to extend our wheeled carts  to more sophisticated mechanical designs to collect human datasets for traversing rough terrains in outdoor environments.

\vspace{8pt}
{\small
\noindent
{\bf Acknowledgements}
We would like to thank Zhenyu Jiang and Braham Snyder for providing feedback on this manuscript.
We acknowledge the support of the National Science Foundation (1955523, 2145283), the Office of Naval Research (N00014-22-1-2204), and Amazon.
}

\renewcommand*{\bibfont}{\footnotesize}
\renewcommand{\baselinestretch}{0.90} 
\printbibliography
\clearpage
\renewcommand{\baselinestretch}{1.0} 

\section* {Appendix}
\label{sec:appendix}

\renewcommand\thesubsection{\Alph{subsection}}

\subsection{Implementation Details}
The navigation controller predicts the target velocity commands at 10Hz. The low-level gait controller takes the buffer of recent velocity commands, robot states and previous joint-space actions as input and produces joint-space action commands at 38Hz to actuate the robot. 
\paragraph{Navigation Controller} 
The navigation policy uses a ResNet18-backbone network~\cite{he2016deep} as the image encoder. The encoded image features are flattened and concatenated with the 1D heading direction value. The concatenated vector is passed through a two-layer multi-layer perceptron (MLP) with 1024 hidden units in each layer. The input RGB-D images have a size of $212\times 120$. For imitation learning, we develop our behavioral cloning implementations with the robomimic framework~\cite{mandlekar2021matters}. For the recurrent neural networks, we use LSTM~\cite{hochreiter1997long} of two layers with 400 hidden units for each layer. The GMM policy output has 5 modes.

\paragraph{Gait Controller} 
Each tuple in the buffer $\mathcal{B}_t$ is a 48D vector, containing $u_i\in\mathbb{R}^2$, $q_i\in\mathbb{R}^{34}$, and $a_{i-1}\in\mathbb{R}^{12}$. $u_i$ consists of the forward linear velocity and angular velocity. $q_i$ consists of joint positions and velocities of each leg (24D), binary foot contact states for each foot (4D), the IMU measurement of the robot's body orientation (3D), and angular velocities (3D). $a_{i-1}$ is a 12D joint-space command of the previous time step.  We choose $T=11$ for the size of the history buffer. We use 1D temporal convolution to transform the input $T+1$ tuples $\mathcal{B}_t$ into a 32D vector. Specifically, the 48D tuple of each time step is first encoded by a linear layer, followed by 3 layers of temporal convolution. The feature from temporal convolution is flattened and projected by another linear layer into the 32D vector.
We concatenate the 48D tuple of the most recent state $(u_t,q_t,a_{t-1})$ and the encoded 32D vector of history states. This concatenated feature is passed through a two-layer MLP of 256 hidden units to produce a Gaussian distribution of joint-space actions. 

The reward function for training $\pi_L$ consists of the terms for tracking commands, balancing the body, minimizing energy consumption, and regulating foot contacts, as below.
\begin{enumerate}
\item Tracking commands:  $K(k_1 |e_{\psi}|^2) e^{k_2 |\boldsymbol{e}_{xy}|} $
\item Balancing the body:  $K(k_3|\boldsymbol{e}_{xy}|^2)K (k_4 |\boldsymbol{\theta}|^2)$
\item Minimizing energy consumption:  $K(k_3|\boldsymbol{e}_{xy}|^2) E$
\item Regulating foot contacts: $k_5 \left(\max(n_1-3, 0) - n_2 \right)$
\end{enumerate}
where $K(\cdot) := 1- \tanh (\cdot) $ and $k_1, ... k_5$ are given as positive weight coefficients. $E$ is the energy consumption, $\boldsymbol{\theta}$ is the vector of roll and pitch,  $\boldsymbol{e}_{xy}$ is the linear velocity errors in the x and y axes, and ${e}_{\psi}$ is the yaw rate error. $n_1$ , $n_2$ are the numbers of foot contacts, and non-foot contacts, respectively.
The terms of tracking commands and balancing the body are designed to improve the stability of tracking commands, and the other two terms improve the gait patterns. 64 actors of distributed PPO are used for training.

Our learned Gait Controller is trained with the reward design prioritizing tracking rapidly changing commands for reactive locomotion skills. Therefore, this reward design yields good command-tracking performance, though the behavior of the Gait Controller is jittery and less energy efficient. The learned Gait Controller outperforms the MPC baseline in terms of tracking time-varying velocity commands, as shown in the simulation unit tests (\textit{Analysis of Gait Controllers}), and the impact of better tracking is discussed in Sec.~\ref{sec:quant-eval}.



\begin{algorithm}[hbt!]
\caption{PRELUDE}
\begin{algorithmic}
\State // Parameters
\State $N$: Number of dataset samples
\State $T$: Buffer length of recent robot states
\State $f_h$: Navigation Controller frequency
\State $f_l$: Gait Controller frequency
\\
\State // Training of PRELUDE
\State $ \mathcal{D} \gets \{(s_i, u_i)\}_{i=1}^N$ \Comment{Human demonstration}
\State $ \pi_H \gets$ trained by Imitation Learning with $ \mathcal{D}$
\State $ \pi_L \gets$ trained by Reinforcement Learning in simulation
\\
\State // Deployment of PRELUDE
\State $ u_t \gets$ initialized \Comment{velocity command}
\State $ a_t \gets$ initialized \Comment{joint-space action} 
\State $\mathcal{B}_t \gets \{(u_i,q_i,a_{i-1})\}_{i=t-T,\ldots,t}$ \Comment{Buffer of recent robot states}
\State $t_h \gets t$ \Comment{Navigation Controller timer}
\State $t_l \gets t$ \Comment{Gait Controller timer}
\While{episode is not done}
  \State $t \gets$ current time
  \State $s_t \gets$ observed by the robot sensors
  \State $q_t \gets$ joint states in $s_t$
  \If{$t \geq t_h$}
   \State $u_t \sim  \pi_H(s_t, u_t)$
   \State $t_h \gets t_h + {1\over{f_h}}$
  \EndIf
  \If{$t \geq t_l$}
    \State Store $(u_t,q_t,a_{t-1})$ in $\mathcal{B}_t$
    \State $a_t \sim \pi_L(\mathcal{B}_t)$
   \State $t_l \gets t_l + {1\over{f_l}}$
  \EndIf
\EndWhile
\end{algorithmic}
\end{algorithm}
\vspace{-6pt}

\subsection{Real Robot Evaluation}

The location and background are unseen in all of the \textbf{Dynamic (Complex)} experiments. Lighting conditions vary for the \textbf{Dynamic (Sparse)} episodes compared with the training data in the same location. This location is sensitive to sunlight, and the resulting background is significantly different from the training data. \textbf{Static (Sparse)} and \textbf{Static (Complex)} are performed over several days in shared spaces; therefore, the background is naturally different. In these two environments, there are 20\% and 50\% unseen objects, respectively. For sparse scenes where objects are large, we were limited in the number of different obstacles available, while in complex scenes, there was more variation. Also, all humans are unseen obstacles as they differ in size, shape, clothing color, and gait style.



\end{document}